# Brain Tumor Segmentation using an Ensemble of 3D U-Nets and Overall Survival Prediction using Radiomic Features

Xue Feng[1][0000-0002-2181-9889] , Nicholas Tustison[2] and Craig Meyer[1, 2]

[1] Biomedical Engineering, University of Virginia, Charlottesville VA 22903, USA
[2] Radiology & Medical Imaging, University of Virginia, Charlottesville VA 22903, USA
xf4j@virginia.edu

**Abstract.** Accurate segmentation of different sub-regions of gliomas including peritumoral edema, necrotic core, enhancing and non-enhancing tumor core from multimodal MRI scans has important clinical relevance in diagnosis, prognosis and treatment of brain tumors. However, due to the highly heterogeneous appearance and shape, segmentation of the sub-regions is very challenging. Recent development using deep learning models has proved its effectiveness in the past several brain segmentation challenges as well as other semantic and medical image segmentation problems. Most models in brain tumor segmentation use a 2D/3D patch to predict the class label for the center voxel and variant patch sizes and scales are used to improve the model performance. However, it has low computation efficiency and also has limited receptive field. U-Net is a widely used network structure for end-to-end segmentation and can be used on the entire image or extracted patches to provide classification labels over the entire input voxels so that it is more efficient and expect to yield better performance with larger input size. Furthermore, instead of picking the best network structure, an ensemble of multiple models, trained on different dataset or different hyper-parameters, can generally improve the segmentation performance. In this study we propose to use an ensemble of 3D U-Nets with different hyper-parameters for brain tumor segmentation. Preliminary results showed effectiveness of this model. In addition, we developed a linear model for survival prediction using extracted imaging and non-imaging features, which, despite the simplicity, can effectively reduce overfitting and regression errors.

## 1 Introduction

Gliomas are the most common primary brain malignancies, with different degrees of aggressiveness, variable prognosis and various heterogeneous histological sub-regions, i.e. peritumoral edema, necrotic core, enhancing and non-enhancing tumor core. This intrinsic heterogeneity of gliomas is also portrayed in their radiographic phenotypes, as their sub-regions are depicted by different intensity profiles disseminated across multimodal MRI (mMRI) scans, reflecting differences in tumor biology. Quantitative analysis of imaging features such as volumetric measures after manual/semi-automatic segmentation of the tumor region has shown advantages in image-based tumor phenotyping over traditionally used clinical measures such as largest anterior-posterior, transverse, and inferior-superior tumor dimensions on a subjectively-chosen slice [1-2]. Such phenotyping may enable assessment of reflected biological processes and assist in surgical and treatment planning. To compare and evaluate different automatic segmentation algorithms, the Multimodal Brain Tumor Segmentation Challenge (BraTS) 2018 was organized using multi-institutional pre-operative MRI scans for the segmentation of intrinsically heterogeneous brain tumor sub-regions [3-4]. More specifically, the dataset used in this challenge includes multiple-institutional clinically-acquired pre-operative multimodal MRI scans of glioblastoma (GBM/HGG) and low-grade glioma (LGG) containing a) native (T1) and b) post-contrast T1-weighted (T1Gd), c) T2-weighted (T2), and d) Fluid Attenuated Inversion Recovery (FLAIR) volumes [5-6]. 285 training volumes with annotated GD-enhancing tumor, peritumoral edema and necrotic and non-enhancing tumor. Furthermore, to pinpoint the clinical relevance of this segmentation task, BraTS'18 also included the task to predict patient overall survival from images together with the patient age and resection status. To tackle these two tasks, this study is performed with two goals: 1) provide pixel-by-pixel label maps for the three sub-regions and background; 2) estimate the survival days.

Convolutional neural network (CNN) based models have proven their effectiveness and superiority over traditional medical image segmentation algorithms and are quickly becoming the mainstream in BraTS challenges. Due to the highly heterogeneous appearance and shape of brain tumors, small patches are usually extracted to predict the class for the center voxel. To improve model performance, multi-scale patches with different receptive field sizes are often used in the model [7]. In contrast, U-Net is a widely used convolutional network structure that consists of a contracting path to capture context and a symmetric expanding path that enables precise localization with 3D extension [8-9]. It can be used on the entire image or extracted patches to provide class labels for all input voxels when padding is used. Furthermore, instead of picking the best network structure, an ensemble of multiple models, trained on different dataset or different hyper-parameters, can generally improve the segmentation performance over a single model due to the averaging effect. In this study we propose to use an ensemble of 3D U-Nets with different hyper-parameters trained on non-uniformly extracted patches for brain tumor segmentation. During testing, a sliding window approach is used to predict class labels with adjustable overlap to improve accuracy. With the segmentation labels, we will develop a linear model for survival prediction using extracted imaging features and additional non-imaging features since the linear models can effectively reduce overfitting and thus regression errors.

## 2 Methods

For the brain tumor segmentation task, the steps in our proposed method include pre-processing of the images, patch extraction, training multiple models using a generic 3D U-Net structure with different hyper-parameters, deployment of each model for full volume prediction and final ensemble modeling. For the survival task, the steps include feature extraction, model fitting, and deployment. Details are described as follows.

### 2.1 Image Pre-processing

To compensate for the MR inhomogeneity, the bias correction algorithm based on N4ITK was applied to the T1, T1Gd images, T2 and flair images [10]. A smooth inhomogeneity field due to variations in coil sensitivity was estimated and compensated from the images. A non-local means denoising method was then used to reduce noise after bias correction [11]. The implementations on ITK [12] were used with a Python wrapper from Nipype [13]. Python-based parallel execution with multiple threads was used to accelerate the two steps. The processed images were stored for future usage. Fig. 1 shows the original T1 image (left), image with only bias correction (center) and image with bias correction and denoising (right). The signal-to-noise ratio (SNR) of the image is increased with the denoising method, which could potentially help improving the segmentation accuracy and robustness against noise.

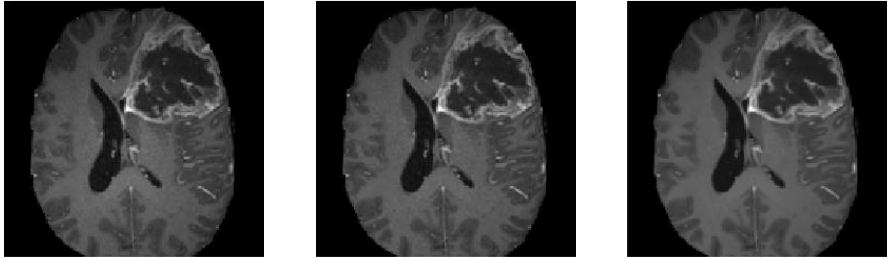

**Fig. 1.** Original T1 image (left), image with only bias correction (center) and image with bias correction and denoising (right). The right image has improved SNR.

As MR images do not have standard pixel intensity values, to reduce the effects from different contrasts and different subjects, each 3D image was normalized to 0 to 1 separately by subtracting the min values and divided by the pixel intensity range. After normalization, for each subject, images of all contrast were fused to form the last dimension so that the whole input image size becomes 155x240x240x4.

## 2.2 Non-uniform Patch Extraction

For simplicity, we will use foreground to denote all tumor pixels and background to denote the rest. There are several challenges in directly using the whole images as the input to a 3D U-Net: 1) the memory of a moderate GPU is often 12 Gb so that in order to fit the model into the GPU, the network needs to greatly reduce the number of features and/or the layers, which often leads to a significant drop in performance as the expressiveness of the network is much reduced; 2) the training time will be greatly prolonged since more voxels contribute to calculation of the gradients at each step and the number of steps cannot be proportionally reduced during optimization; 3) as the background voxels dominate the whole image, the class imbalance will cause the model to focus on background if trained with uniform loss, or prone to false positives if trained with weighted loss that favors the foreground voxels. Therefore, to more effectively utilize the training data, smaller patches were extracted from each subject. As the foreground labels contain much more variability and are the main targets to segment, more patches from the foreground voxels should be extracted.

In implementation, during each epoch, a random patch was extracted from each subject using non-uniform probabilities. The valid patch centers were first calculated by removing edges to make sure each extracted patch was completely within the whole image. The probability of each valid patch center $p_{i,j,k}$ was calculated using the following equation:

$$p_{i,j,k} = \frac{s_{i,j,k}}{\Sigma_{i,j,k} s_{i,j,k}} \qquad [1]$$

in which $s_{i,j,k} = 1$ for all voxels with maximal intensity lower than the $1^{st}$ percentile, $s_{i,j,k} = 6$ for all foreground voxels and $s_{i,j,k} = 3$ for the rest. The patch center was then randomly selected based on the calculated probability and the corresponding patch was extracted. Since normal brain images are symmetric along the left-right direction, a random flip along this direction was made after patch extraction. No other augmentation was applied.

Before training, the per-input-channel mean and standard deviation of extracted patches were calculated by running the extraction process 400 times, with each time using a randomly selected training subject. The extracted patches were then subtracted with the mean and divided by the standard deviation along each input channel.

## 2.3 Network Structure and Training

A 3D U-Net based network was used as the general structure, as shown in Fig. 2. Zero padding was used to make sure the spatial dimension of the output is the same with the input. For each encoding block, a VGG like network with two consecutive 3D convolutional layers with kernel size 3 followed by the activation function and batch norm layers were used. The parametric rectilinear function (PReLU), given as:

$$f(x) = \max(0, x) - \alpha \max(0, -x) \qquad [1]$$

was used with trainable parameter $\alpha$ as the activaton function. The number of features was doubled while the spatial dimension was halved with every encoding block, as in conventional U-Net structure. To improve the expressiveness of the network, a large number of features were used in the first encoding block. Dropout with ratio 0.5 was added after the last encoding block. Symmetric decoding blocks were used with skip-connections from corresponding encoding blocks. Features were concatenated to the de-convolution outputs. The extracted segmentation map of the input patch was expanded to the multi-class the ground truth labels (3 foreground classes and the background). Weighted/non-weighted cross entropy was used as the loss function.

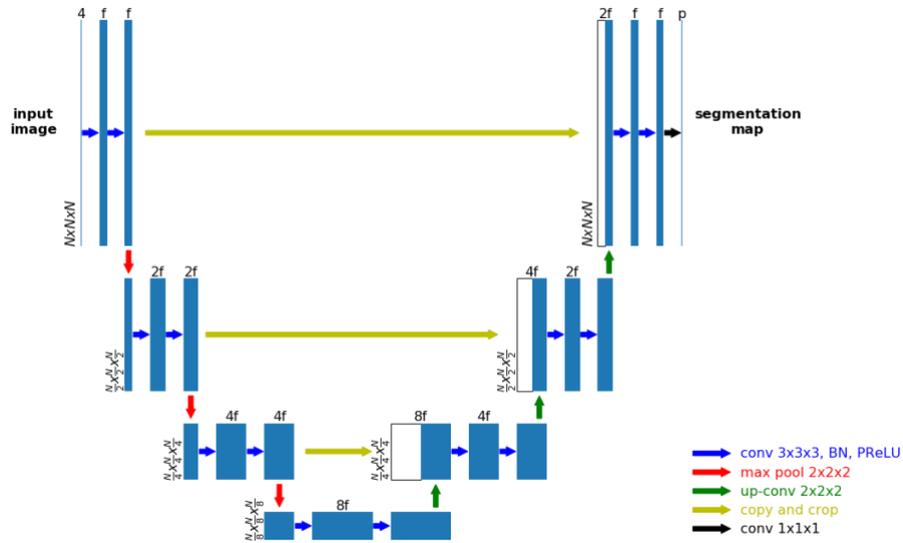

**Fig. 2.** 3D U-Net structure with 3 encoding and 3 decoding blocks.

The number of encoding/decoding blocks, the weights in the loss function and the patch size were chosen as the tunable hyper-parameters when constructing multiple models. Due to memory limitations, for a larger patch size, the number of features needs to be reduced. In current implementation, due to constraint in computational resources, six models were trained, with detailed parameters shown in Table 1. N denotes the input size, M denotes the number of encoding/decoding blocks and f denotes the input features at the first layer. For weighted loss, 1.0 was used for background and 2.0 was used for each class of foreground voxels.

**Table 1.** Detailed parameters for all 6 3D U-Net models.

| Model # | M | N | f | Loss Type |
|---------|---|-----|----|-----------|
| 1 | 3 | 64 | 96 | Uniform |
| 2 | 3 | 64 | 96 | Weighted |
| 3 | 4 | 64 | 96 | Uniform |
| 4 | 4 | 96 | 96 | Weighted |
| 5 | 3 | 80 | 64 | Uniform |

| 6 | 3 | 80 | 64 | Weighted |
|---|---|---|---|---|

Training was performed on a Nvidia Titan Xp GPU with 12 Gb memory. 640 epochs were used. As mentioned earlier, during each epoch, only one patch was extracted every subject. Subject orders were randomly permuted every epoch. The Tensorflow framework was used with Adam optimizer. Batch size was set to 1 during training. During testing, as a smaller batch size was very sensitive to the running statistics, all batch norm layers did not use the running statistics but the statistics of the batch itself. This is usually called a layer normalization as it normalizes each feature map with its own mean and standard deviation. A learning rate of 0.0005 was used without further adjustments during training. The total training time was about 60 hours.

## 2.4 Volume Prediction Using Each Model

Due to the fact that the entire image cannot fit into the memory during deployment, a sliding window approach needs to be used to get the output for each subject. However, as significant padding was made to generate the output label map at the same size as the input, boundary voxels of a patch were expected to yield unstable predictions when sliding the window across the whole image without overlaps. To alleviate this problem, a stride size at a fraction of the window size was used and the output probability was averaged. In implementation, the deployment window size was chosen to be the same as the training window size, and the stride was chosen as ½ of the window size. For each window, the original image and left-right flipped image were both predicted, and the average probability after flipping back the output of the flipped input was used as the output. Therefore, each voxel, except for a few on the edge, will be predicted 16 times when sliding across all directions. Although smaller stride sizes can be used to further improve the accuracy with more averages, the deployment time will be increased 8 times for every ½ reduction of the window size and thus will quickly become unmanageable. Using the parameters as mentioned on the same GPU, it took about 1 minutes to generate the output for the entire volume per subject. Instead of performing a thresholding on the probability output to get the final labels, the direct probability output was saved for each model to the disk.

## 2.5 Ensemble Modeling

The ensemble modeling process was rather straightforward. The probability output of all classes from each model was read from the disk and the final probability was calculated via simple averaging. The class with the highest probability was selected as the final segmentation label of each voxel.

## 2.6 Survival Prediction

To predict the post-surgery survival time measured in days, extracted images features and non-image features were used to construct a linear regression model. 6 image features were calculated from the ground truth label maps during training and the predicted label maps during validation. For each foreground class, the volume (V) by summing up the voxels and the surface area (S) by summing up the magnitude of the gradients along three directions were obtained, as described in the following equations

$$V_{ROI} = \sum_{i,j,k} s_{i,j,k} \qquad [2]$$

$$S_{ROI} = \sum_{i,j,k} s_{i,j,k} \sqrt{(\frac{\partial s}{\partial i})^2 + (\frac{\partial s}{\partial j})^2 + (\frac{\partial s}{\partial k})^2} \qquad [3]$$

in which ROI denotes a specific foreground class and $s_{i,j,k} = 1$ for voxels that are classified to belong to this ROI and $s_{i,j,k} = 0$ otherwise.

Age and resection status were used as non-imaging clinical features. As there were two classes of resection status and many missing values of this status, a two-dimensional feature vector was used to represent the status, given as GTR: (1, 0), STR: (0, 1) and NA: (0, 0). A linear regression model after normalizing the input features to zero mean and unit standard deviation was fit with the training data. As the input feature size is 9, the risk for overfitting is greatly reduced.

## 3    Results

### 3.1    Brain Tumor Segmentation

All 285 training subjects were used in the training process. 66 subjects were provided as validation. The dice indexes, sensitivities and specificities, 95 Hausdorff distances of the enhanced tumor (ET), whole tumor (WT) and tumor core (TC) were automatically calculated after submitting to the CBICA's Image Processing Portal. With multiple submissions, we were able to compare the performances of each individual model and the final ensemble.

**Table 2.** Performances of each individual model and the ensemble

| Model # | Dice_ET | Dice_WT | Dice_TC | Dist_ET | Dist_WT | Dist_TC |
|---------|---------|---------|---------|---------|---------|---------|
| 1 | 0.7688 | 0.9015 | 0.8237 | 4.1270 | 4.5437 | **5.5226** |
| 2 | 0.7677 | 0.9066 | 0.8248 | 4.2218 | 6.4637 | 8.8593 |
| 3 | 0.7695 | 0.9040 | 0.8306 | 7.1372 | 8.9214 | 11.4460 |
| 4 | 0.7707 | 0.8990 | 0.8104 | **3.1454** | 6.0081 | 6.9814 |
| 5 | 0.7863 | 0.9078 | 0.8217 | 4.1894 | 4.5704 | 6.2030 |
| 6 | 0.7616 | 0.8917 | 0.8149 | 4.2222 | 4.1053 | 6.9598 |
| Ensemble | **0.7917** | **0.9094** | **0.8362** | 4.0186 | **3.8009** | 5.6451 |

Table 2 shows the mean dice scores and 95 Hausdorff distances of ET, WT and TC for the 6 individual models and the ensemble of them. Sensitivity and specificity are highly correlated with the dice indexes so that they are not included. The best performance of each evaluation metric is highlighted. All 3D U-Net models perform similar but the ensemble of them has the overall best performances as compared with each individual model. It is also noticed that weighted cross-entropy loss has high sensitivity but lower specificity compared with the uniform counterpart, which is likely due to the fact that by assigning more weights to the foreground, the network tends to be more aggressive in assigning foreground labels.

### 3.2 Survival Prediction

All 163 training subjects with survival data were used in the training process. The training coefficient of determination was 0.259. 28 cases were evaluated after submitting to the CBICA's Image Processing Portal. The accuracy was 0.321, MSE was 99115.86, median SE was 77757.86, std SE was 104291.596 and Spearman Coefficient was 0.264. The performance on the validation dataset is not as accurate as other top teams in this task, however, our method won the 1[st] place in the testing dataset, which is likely due to significant overfitting of other teams in validation. The final result is encouraging and shows that a linear model is robust against overfitting.

## 4 Discussion and Conclusions

In this paper we developed a brain tumor segmentation method using an ensemble of 3D U-Nets. Bias correction and denoising were used as pre-processing. 6 networks were trained with different number of encoding/decoding blocks, input patch sizes and different weights for loss. The preliminary results showed an improvement with ensemble modeling. For survival prediction, we used a simple linear regression by combining radiomics features from images such as volumes and surface areas of each sub-region and non-imaging clinical features.

For segmentation, it is noted that the median metrics are significantly higher than the mean metrics. For example, the median dice indexes were 0.867, 0.923 and 0.904 for ET, WT and TC in the final ensembled model. It makes sense in that the theoretical maximum dice index is 1 and minimum dice index is 0. However, we noted that in several cases, the dice indexes are as low as 0 for ET and TC and 0.6 for WT. It is mostly due to the low sensitivity meaning that the model is not able to recognize the corresponding tumor regions. The possible reason for these failed regions is that their characteristics deviate a lot from the training dataset. This is also encouraging in that for majority of the cases, the segmentation quality is very high.

In the 3D U-Net model, we found that the batch norm layer was helpful in improving the model stability and performance. However, different with the canonical application of the batch norm layer, in which the batch statistics is used in training and the global statistics is used in deployment, it performed much better with batch statistics in deployment than global statistics. Since the batch size is 1, a per-channel normalization is actually performed by subtracting its own mean. One possible explanation could be that by doing such normalization, the model focuses on the differences of neighboring pixels in one channel and ignores the absolute values, which may help the segmentation process. However, further investigation is needed to figure out the exact reason.

Compared with the patch-based model that only predicts the center pixel, when predicting the segmentation label maps for the full patch, different pixels are very likely to have different effective receptive field sizes due to the zero padding in the edge. We argue that a pixel should still be able to be predicted even based on partial receptive field, which, for the very edge pixel, corresponds to only half of the maximal receptive field. Furthermore, the significant overlap in the sliding windows during deployment can improve the accuracy with more averages.

In the current implementation, 6 networks were trained due to limitations in computation time. It is expected with more networks, the results can be further improved, although the marginal improvement is expected to decrease.

For the survival prediction task, since it is very likely to overfit with such a small dataset and we argue that as many other features may play more important roles in overall survival such as histological and genetic features but unfortunately, they are not available in this challenge, a linear regression model was the safest option to minimize the test errors, although at the cost of its expressiveness. Further exploration of those additional features through clinical collaboration is expected to improve the accuracy of survival prediction.

In conclusion, we developed an ensemble of 3D U-Nets for brain tumor segmentation. The network hyper-parameters are varied to obtain multiple trained models. A linear regression model was also developed for the survival prediction task. Our survival prediction model won the 1st place in the final stage of the competition. The code is available at https://github.com/xf4j/brats18. The paper that summarizes the challenge is available at [14].